# Object Tracking in Satellite Videos Based on a Multi-Frame Optical Flow Tracker

Bo Du, *Senior Member, IEEE*, Shihan Cai, Chen Wu, *Member, IEEE*, Liangpei Zhang, *Senior Member, IEEE*, and Dacheng Tao, *Fellow, IEEE*

*Abstract*—Object tracking is a hot topic in computer vision. Thanks to the booming of the very high resolution (VHR) remote sensing techniques, it is now possible to track targets of interests in satellite videos. However, since the targets in the satellite videos are usually too small compared with the entire image, and too similar with the background, most state-of-the-art algorithms failed to track the target in satellite videos with a satisfactory accuracy. Due to the fact that optical flow shows the great potential to detect even the slight movement of the targets, we proposed a multi-frame optical flow tracker (MOFT) for object tracking in satellite videos. The Lucas-Kanade optical flow method was fused with the HSV color system and integral image to track the targets in the satellite videos, while multi-frame difference method was utilized in the optical flow tracker for a better interpretation. The experiments with three VHR remote sensing satellite video datasets indicate that compared with state-of-the-art object tracking algorithms, the proposed method can track the target more accurately.

## I. INTRODUCTION

OBJECT tracking is one of the most important topics in computer vision. It has been successfully used in many fields, such as surveillance, human-computer interactions and medical imaging [1-4]. Given the initial position and extent of a target object in the initial image, the goal of object tracking is to estimate the position and extent of the target in subsequent frames. Recently, significant efforts have been made in improving object tracking technology [5-12]. Object tracking includes discriminative methods, correlative filter methods, deep learning methods etc. While discriminative model has been widely adopted in object tracking [5-7, 13], algorithms that extract color features [14-17] have also caught many attractions recently. In recent years, the correlation filter-based trackers [2, 18-23] are popular due to their surprising high-speed computing capability and accuracy; besides, neural convolution network(CNN) applying in object tracking have obtain quite satisfying results [23-27].

Recently, commercial satellite technology has achieved significant development in using remote sensing devices to capture very high resolution(VHR) spaceborne videos[28]. Satellite video can acquire a period of continuous observation over a certain area. With satellite videos[29], we can acquire more dynamic information such as the moving trajectory, speed and directions of a target object, which are unavailable in traditional remote sensing satellite static images[30]. In 2013, Skybox Imaging launched the first commercial satellite, SkySat-1, which opened a new chapter of satellite video sensor for providing video data with a resolution of one meter[31]. Then, UrtheCast installed the high-resolution camera, Irish, on the International Space Station (ISS) in 2013. On the other hand, China launched "Jilin-1" commercial satellite in 2015, and launched Jilin-1 Agile Video Satellites in 2017. With the launch of these commercial video satellites, applying object tracking technologies in satellite video data has been possible. Satellite video data provide great potentials in motion analysis [32], traffic monitoring [33-35], suspicious object surveillance [36], and urban management etc.. Therefore, it shows great significance in studying object tracking technology on satellite video data.

However, satellite video tracking has confronted some problems compared with traditional object tracking task. The difficulties of applying object tracking technology in satellite video data include: 1) Larger scene size. The width and height of satellite video are hundreds of times larger than traditional object tracking videos. It results in full image searching more difficult, and time-consuming; 2) Smaller target size. In satellite videos, the interested target only takes up about 0.01‰ of full video frame pixels or even less, resulting in various problems in object detection and tracking; 3) Less features and more similar background. The small target contains less features, and is more similar with the background, which has obviously lower

This paragraph of the first footnote will contain the date on which you submitted your paper for review.

B. Du is with the School of Computer, and Collaborative Innovation Center of Geospatial Technology, Wuhan University, Wuhan, P.R. China (e-mail: gunspace@163.com).

S. Cai is with the School of Computer, Wuhan University, Wuhan, P.R. China (e-mail: twilight@whu.edu.cn).

C. Wu is with the State Key Laboratory of Information Engineering in Surveying, Mapping, and Remote Sensing, and School of Computer, Wuhan University, Wuhan, P.R. China (e-mail: chen.wu@whu.edu.cn).

L. Zhang is with the Remote Sensing Group, State Key Laboratory of Information Engineering in Surveying, Mapping, and Remote Sensing, Wuhan University, Wuhan, P.R. China (e-mail: zlp62@whu.edu.cn).

D. Tao is with the School of Information Technologies and the Faculty of Engineering and Information Technologies, and the UBTECH Sydney Artificial Intelligence Centre, the University of Sydney, Sydney, Australia(e-mail: dacheng.tao@sydney.edu.au)

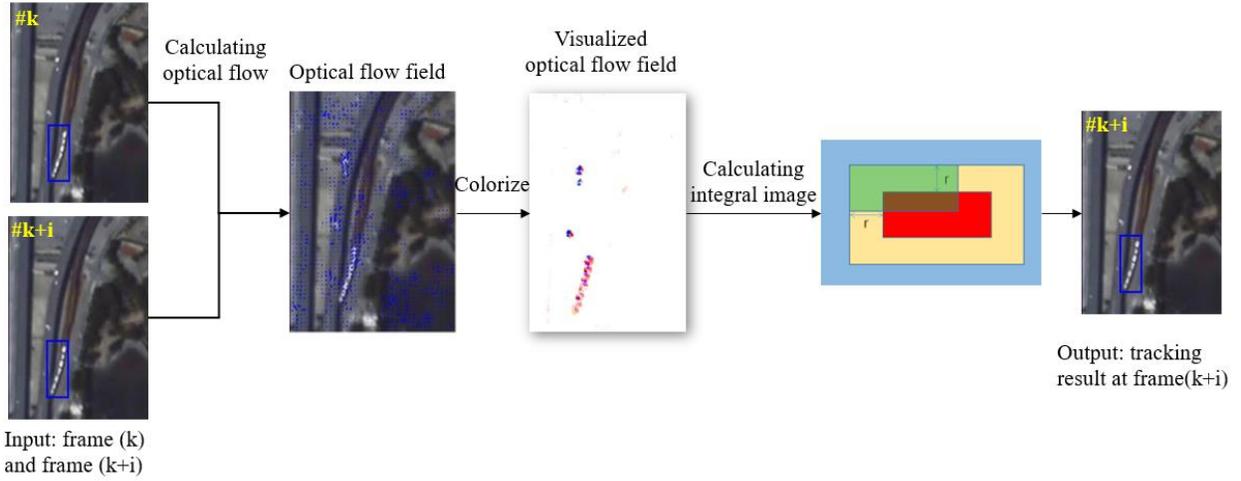

Fig. 1 Procedure of the proposed MOFT method.

resolution than traditional object tracking data; 4) Bigger illumination variation influence. As the satellite video data is taken from high altitude, illumination variation attribute has greater impact in object tracking performance.

In order to solve the aforementioned problems, it is a good solution to employ optical flow. Optical flow is the apparent motion of the brightness patterns in the image, which can provide important information for the small motion of an object [37]. It is worth noting that in satellite video, the background surrounding the interested target mostly keep unchanged, thus optical flow can achieve good performance in separating the target and background. Besides, if objects move too slow to be analyzed for optical flow, multi-frame difference can be employed to improve the tracking performance [38].

Therefore, in this paper, we proposed a multi-frame optical flow tracker (MOFT) to track moving objects in satellite videos. The satellite video frames are firstly disposed by Lucas-Kanade optical flow method to get an optical flow field. Then, the HSV Color system is used for converting two-dimensional optical flow field into a three-channel color image. And finally, the integral image is employed to obtain the most probable position of target. In addition, since most interested targets move slightly in satellite videos, multi-frame difference method is applied to locate a more accurate position of the moving target.

The rest of this paper is organized as follows. Section II details the proposed object tracking method. The experiments and discussions are presented in Section III. Finally, the conclusion is drawn in Section IV.

## II. METHODOLOGY

In this section, we elaborate how to improve optical flow method with HSV Color system and integral image to track the interested target in satellite video data. The whole procedure of the proposed method is shown in Fig. 1.

The main steps are as follows:

1) Input two frames of satellite video data into optical flow method to obtain the optical flow field;

2) Employ HSV Color system to convert the two-dimension optical flow field, which include the vector information $(x, y)$, into three-bands RGB image;

3) Employ rectangle integral box, which has the same size as the ground truth bounding box, to integral RGB image to get an integral matrix;

4) Find the minimum value inside the integral matrix, and the relative position for the minimum value is the central location of the target;

5) Implement the location of the target to get the tracked bounding box.

### A. Lucas-Kanade Optical Flow

Optical flow is the apparent motion of the brightness patterns in the image, and the motion field is projected from three-dimension motion into two-dimension plane [37, 39]. Optical flow algorithm estimates two-dimensional motion vector for each pixel between two frames. Lucas-Kanade Optical Flow [40] conforms to Brightness Constancy principle, which is one of the basic principle of optical flow. The Brightness Constancy principle assumes that when a pixel flows from one image to another, its intensity or color does not change. Thus, if $I(x, y, t)$ is the intensity of a pixel $(x, y)$ at time t, $dx$ is the x-direction displace distance, $dy$ is the y-direction displace distance, *Brightness Constancy* can be written as:

$$I(x, y, t) = I(x + dx, y + dy, t + 1) \quad (1)$$

Linearizing Eq.(1) by applying a first-order Taylor expansion to the right-hand side, we can get Eq.(2):

$$I(x, y, t) = I(x, y, t) + \frac{\partial I}{\partial x} dx + \frac{\partial I}{\partial y} dy + \frac{\partial I}{\partial t} dt + \epsilon \quad (2)$$

where $\epsilon$ contains second and higher order terms in $dx$, $dy$, and $dt$. After subtracting $I(x, y, t)$ from both sides and dividing through by $dt$, we can simplify Eq.(2) to get the *Optical Flow Constraint equation*:

$$\frac{\partial I}{\partial x}\frac{dx}{dt} + \frac{\partial I}{\partial y}\frac{dy}{dt} + \frac{\partial I}{\partial t} = 0 \quad (3)$$

Let

$$v_x = \frac{dx}{dt} \text{ and } v_y = \frac{dy}{dt} \quad (4)$$

where $v_x$, $v_y$ is the optical flow vector's *x, y* directional component.

Then Eq.(3) can be written as

$$I_x v_x + I_y v_y = -I_t \quad (5)$$

Both the Brightness Constancy and Optical Flow Constraint equation provide just one constraint on the two variables for each pixel. This is the origin of the optical flow method's Aperture Problem. In order to solve this problem, Lucas-Kanade algorithm is applied.

According to the optical flow's Spatial Coherence assumption, neighboring points in same surface of three-dimension have similar motion, and project to nearby points on the two-dimensional image plane [41]. Therefore, we can work out the velocity value of the central pixel by using the surrounding pixels to set up a group of equations. Assume the optical flow velocity $v$ ($v_x$, $v_y$) in a $m \times m$ ($m = 2 \times n + 1, n > 0$) window is a constant, then we can get a group of equations from pixels $i$, $i = 1, \dots, n, n = m^2$, as follows:

$$\begin{cases} I_{x_1}v_x + I_{y_1}v_y = -I_{t_1} \\ I_{x_2}v_x + I_{y_2}v_y = -I_{t_2} \\ \vdots \\ I_{x_n}v_x + I_{y_n}v_y = -I_{t_n} \end{cases} \quad (6)$$

which can be expressed as:

$$\begin{bmatrix} I_{x_1} & I_{y_1} \\ I_{x_2} & I_{y_2} \\ \vdots & \vdots \\ I_{x_n} & I_{y_n} \end{bmatrix} \begin{bmatrix} v_x \\ v_y \end{bmatrix} = \begin{bmatrix} -I_{t_1} \\ -I_{t_2} \\ \vdots \\ -I_{t_n} \end{bmatrix} \quad (7)$$

Now we have an overconstrained system, and this system can be solved if it contains more than one point that are included in this window. We can simplify Eq.(7) as

$$Ad = b \quad (8)$$

To work out this system, we set up a least-squares minimization of the equation, whereby $\min\|Ad - b\|^2$ is solved in standard from as:

$$(A^T A)d = A^T b \quad (9)$$

Writing this out in more detail:

$$\begin{bmatrix} \sum I_x I_x & \sum I_x I_y \\ \sum I_x I_y & \sum I_y I_y \end{bmatrix} \begin{bmatrix} v_x \\ v_y \end{bmatrix} = -\begin{bmatrix} \sum I_x I_t \\ \sum I_y I_t \end{bmatrix} \quad (10)$$

From this relation we can obtain the $v_x$ and $v_y$ motion components. The solution to this equation is then:

$$\begin{bmatrix} v_x \\ v_y \end{bmatrix} = -\begin{bmatrix} \sum I_x I_x & \sum I_x I_y \\ \sum I_x I_y & \sum I_y I_y \end{bmatrix}^{-1} \begin{bmatrix} \sum I_x I_t \\ \sum I_y I_t \end{bmatrix} \quad (11)$$

Now, we have gotten a two-dimension optical flow field which has same width and height with the original image. However, in the obtained optical flow field, the velocity of each pixel's optical flow has different directions, which makes the analysis impossible. To address this problem, we introduce HSV color system to turn the pixel's optical flow velocity into three bands RGB color.

*B. HSV color system*

There are several color spaces in color representation, and each of them has its own advantages and disadvantages. The RGB color space can be geometrically represented in a three dimensional cube [42], while the optical flow vector we calculated is represented only by two dimensions $(x, y)$, which can form an angle range from 0° to 360°. Thus, the RGB color space is not suitable to represent the optical flow vectors. Instead, using HSV color model to describe the optical vectors is more appropriately.

HSV is one of the most commonly used color space in computer vision [42], since it is more suitable than the colorcube to present radial information [43]. HSV model can describe a color in three dimensions: hue, saturation, and value, which can be geometrically described as cylindrical.

Hue represents basic colors, and is determined by the light wavelengths in the spectrum. Hue can be considered as an angle between a reference line and the color point in color space. The range of the hue value is from 0° to 360°[42].

Saturation measures the radial distance from the cylinder center [42], is the colorfulness of a color [44].

Value measures the departure of a hue from black[43], is the brightness variable of a color.

We implement HSV representation by using the sophisticated approaches that proposed by Zimmer et al [45] and realized by Baker et al[39]. Fig. 2 is an example of optical flow field transformed by HSV color system.

As Fig. 2c shows, after the HSV color system converted the two-dimension optical flow velocity into the three bands RGB color, the target we interested was separated from the complicated background. Since approximate position of the target have been highlighted, we can employ integral image to find the location that has the lowest regional color value integral value.

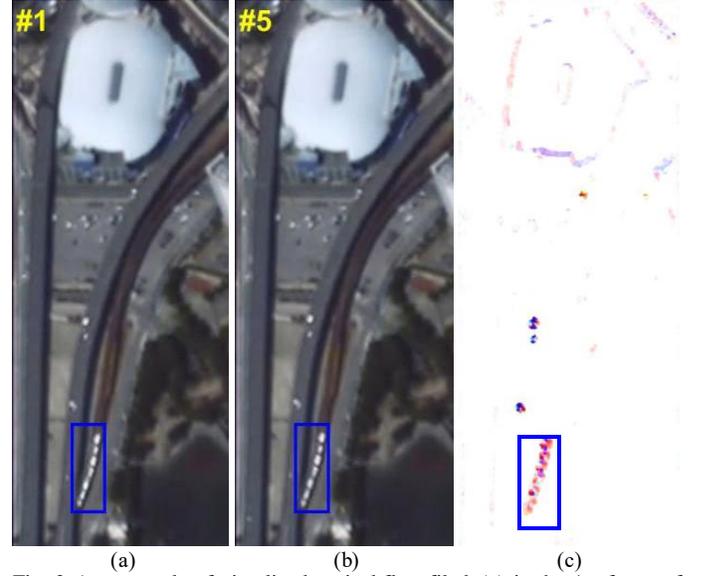

(a)   (b)   (c)

Fig. 2 An example of visualized optical flow filed. (a) is the 1st frame of Vancouver data set, (b) is the 5th frame of Vancouver data set, (c) is the visualized result of optical flow filed of the first frame and fifth frame. The target is the train in blue rectangle.

*C. Integral Image*

The integral image is first introduced in 1984 [46] and applied in computer vision in 1995 [47]. It is an algorithm for quickly and efficiently generating the sum of values in a rectangular subset of a grid [48]. It can be computed from an image using a few operations per pixel. Once computed, a regional integral value can be computed at any scale or location in a constant time [49, 50].

The integral image at location $(x, y)$ contains the sum of the pixels above and to the left of $x, y$, inclusive:

$$ii(x, y) = \sum_{x' \leq x, y' \leq y} i(x', y') \quad (12)$$

where $ii(x, y)$ is the integral image and $i(x, y)$ is the value of the original image.

Using the following pair of recurrences:
$$s(x, y) = s(x - 1, y) + i(x, y) \quad (13)$$
$$ii(x, y) = ii(x, y - 1) + s(x, y) \quad (14)$$

where $s(x, y)$ is the cumulative row sum, and can be initialized as:
$$s(-1, y) = 0 \quad (15)$$
$$ii(x, -1) = 0 \quad (16)$$

We can get some equations from above:
$$s(0, y) = i(0, y) \quad (17)$$
$$ii(x, 0) = s(x, 0) \quad (18)$$

In one-band image, the integral image can be computed in one pass over the original image. Since there are three bands in a RGB image, we should integral the image band-by-band.

As the Fig. 2 shows, we can find that, when the pixel moves more quickly, the optical flow velocity value will be larger. As the optical flow velocity value become larger, one band of RGB color grey level will be higher while the other two bands grey level are lower. just as Fig. 3 shows. Therefore, the region, that is near the target and has the lowest integral result, is the most probable location of our target.

Since we know that the target in satellite videos moves slowly between two frames, there is no need to integral the full frame to find the best tracking location. We can confine the integral place to a rectangular search area that just expand $r$ pixels out of the tracked bounding box of previous frame, where $r$ is the searching radius. As the search area is limited, the computing time will be reduced.

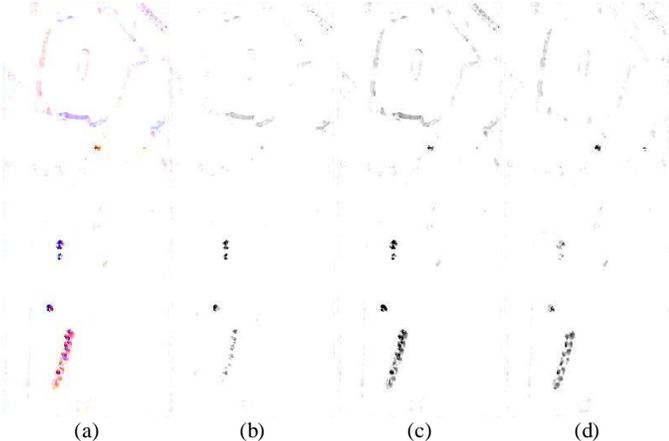

(a) (b) (c) (d)
Fig. 3 An example of visualized optical flow filed and the channel images. (a) is the 1st frame and the 5th frame of Vancouver data set visualized optical flow field image, (b), (c), (d) is visualized optical flow field's red, green, blue channel image, respectively.

### D. Multi-Frame Difference

Because some targets in satellite videos move very slightly, we employ multi-frame difference method to get a more obvious result. As shown in Fig. 4, it can be found that when calculated with multi-frame difference instead of two neighboring frames, the moving target can be better highlighted in the optical flow field (shown in Fig. 4(a) and (b)). However, when the interval between two frames become too large (as shown in Fig. 4 (c)), the optical flow field become so messy that the accurate location of the interested target may be lost and falsely detected. Therefore, it is also very important to find the best interval between two frames to get the optimum performance. The best interval $i$ was evaluated in the experiments.

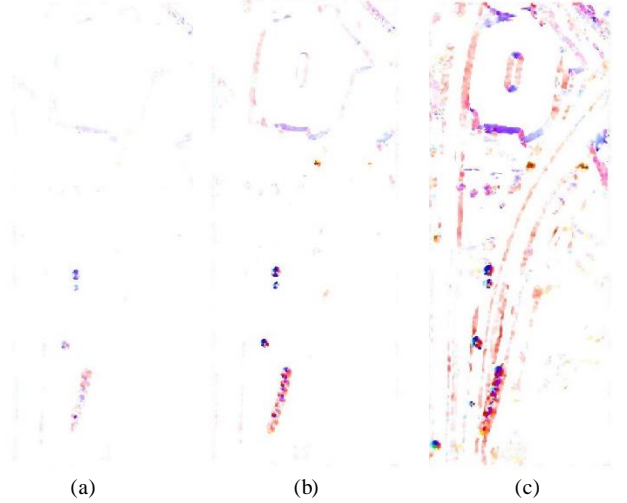

(a) (b) (c)
Fig. 4 visualized optical flow field of Vancouver data set, which are (a) the visualized optical flow field of Vancouver data set between the 1st frame and the 2nd frame; (b) the visualized optical flow field of Vancouver data set between the 1st frame and the 5th frame; (c) the visualized optical flow field of Vancouver data set between the 1st frame and the 10th frame.

The basic steps of our algorithm are presented as Algorithm 1.

**Algorithm 1** Procedure of the proposed MOFT method

**Input:** image of frame $(k)$ and frame $(k + i)$, target tracked bounding box position $P_k(x, y)$ of frame k, search radius r.

**Method:**
a) Calculate the optical flow field between frame $(k)$ and frame $(k + i)$
b) Employ HSV color system to convert the optical flow velocity into RGB color bands.
c) Employ integral box to integral RGB image in search area, which is $r$ pixels out of $P_k(x, y)$, get an integral matrix $I(x, y)$.
d) Find the minimum in $I(x, y)$, regard the $x, y$ position of the minimum as the tracked bounding box result $P_{k+i}(x, y)$ of frame $(k + 1)$.

**Output:** target tracked bounding box position $P_{k+i}(x, y)$ at frame $(k + i)$

## III. EXPERIMENTS

Three videos are used in the experiments, provided by the UrtheCast Corp. and the Chang Guang Satellite Technology Co.,Ltd. respectively. The first satellite video covered Canada Vancouver harbor. The second video covered India New Delhi.

The third video covered Afghanistan Kabul. Fig. 5 shows the first frame of these three data sets. The tracking targets are trains in Vancouver and New Delhi data, and plane in Kabul data.

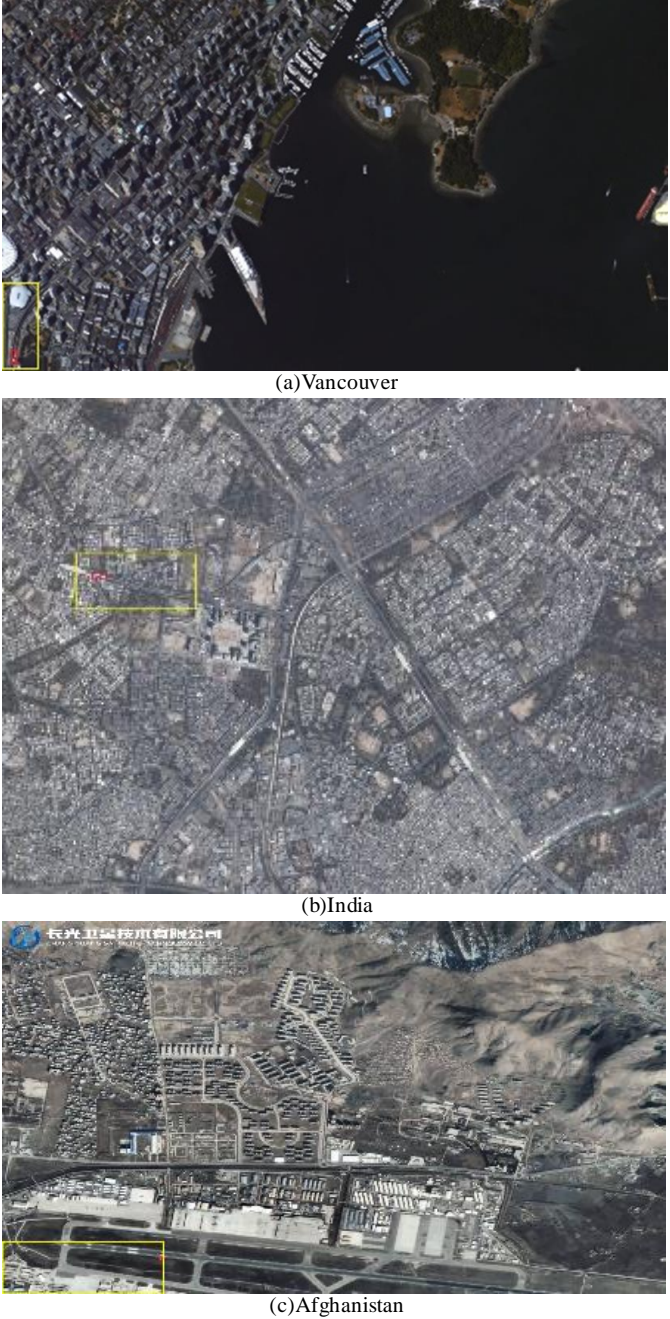

(a)Vancouver

(b)India

(c)Afghanistan

Fig. 5 The 1st frame of our experimental data sets. The target is in the red rectangle and the yellow rectangle indicates our experimental crop area.

As Fig. 5 shown, since the target is so small in the frame image that it will be hard to see the tracking target and tracked bounding box, the video frames are cropped to get a smaller area relatively close to the moving target. The experimental area of this three data sets are shown in Fig. 6. Besides, we initialized the target position in the 1st frame, and evaluated the proposed algorithm by comparing the tracked bounding box with the ground truth bounding box. For comparison, 8 state-of-the-art tracking algorithms are employed, which are Meanshift [51, 52], CT [5], TLD [6], Struck[7, 53], KCF [19], SAMF [21], fDSST [20], and Staple[17]. The compared algorithm are all state-of-the-art methods, and some of them are proposed in recent years and show the best performances in the remarkable datasets[54-59].

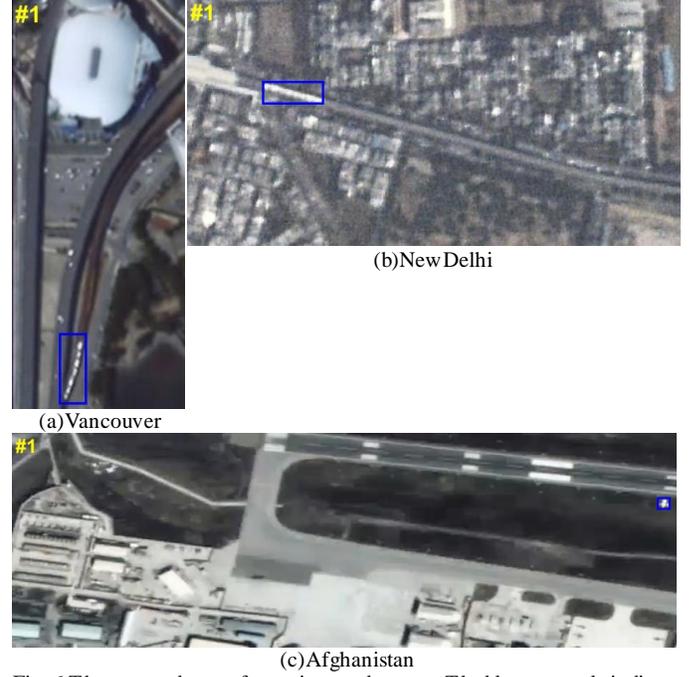

(b)New Delhi

(a)Vancouver

(c)Afghanistan

Fig. 6 The cropped area of experiments data sets. The blue rectangle indicates the object we tracked.

The proposed algorithm is implemented by the Mixture of Matlab and C/C++ on 32 GB memory with Intel Core i7-3770 CPU(3.4GHz). The running speeds of the proposed algorithm are shown in Table I. The speed is mainly affected by the calculation of optical flow field, which is very time-consuming. Since the calculation time may be influenced by hardware and software, we repeat each experiment 100 times to take the average frame per second (FPS). It can be seen that the speeds of the proposed MOFT algorithm are all higher than 20 FPS, and can reach 35 FPS in Vancouver data where the cropped area of this dataset is the smallest in these three experiment datasets. With the code optimization and the improvement of hardware, it is not hard for the proposed method to reach the tracking speed of 30 FPS, and become a real-time tracker.

Table I
FPS of the proposed method in each experiment

| Data set | Vancouver | New Delhi | Kabul |
|---|---|---|---|
| Mean FPS | 35.2 | 20.8 | 20.5 |

As for assessment metrics, the success plot and the precision plot are adopted in the experiments [1, 46].

In success plot, Given the tracked bounding box $r_t$ and the ground truth bounding box $r_g$, the overlap score is defined as

$$S = \frac{|r_t \cap r_g|}{|r_t \cup r_g|} \qquad (19)$$

where ∩ and ∪ represent the intersection and union of two regions, respectively, and $|\cdot|$ denotes the number of pixels in the region. To measure the performance of a sequence, we count the number of successful frames whose overlap $S$ is larger than the given threshold $t_o$. We use the area under curve (AUC) of each success plot to rank the tracking algorithms.

In precision plot, we employ center location error (CLE), which is defined as the average Euclidean distance between the center locations of the tracked bounding box and the manually labeled ground truth [1]. To measure the performance on the sequence of frames, we count the number of successful frames whose CLE is smaller than the given threshold $t_d$.

For assessment, since the AUC score of success plot measures an overall performance, it is adopted to rank the performance of algorithms. Besides, since the metrics of success plots and precision plots is different, the rankings in the success plots and the precision plots may be different. Compared with the center location of the target, the target area is more interested in tracking task, thus we mainly analyze the rankings based on success plots and use the precision plots as an auxiliary.

### A. Canada Vancouver harbor

The first data set is a full color, ultra high definition(UHD) MPEG-4 file that has a spatial resolution of one meter, provided for the 2016 IEEE GRSS Data Fusion Contest by Deimos Imaging and Urthecast, acquired from the International Space Station (ISS)'s High-Resolution camera, Irish, on July 2nd, 2015. The dataset last 34 seconds, having 418 frames, the frame size is $3840 \times 2160$, covering an urban and harbor area in Vancouver, Canada, with the area of about 3.8km $\times$ 2.1km. It mainly describes the traffic situation of this area. The cropped area is from original frame's coordinate (0,1650), experimental area size is $200 \times 500$. The target we tracked is a train, and ground truth bounding box size is $30 \times 80$.

For the parameter interval $i$ and the searching radius $r$, we use experiments to find the best pair.

For the searching radius $r$, the success plot AUC scores are listed in Table II, the best result is in red. It can be seen in Table II, the result with the search radius of 5 can get the best accuracy. However, the other radiuses can also lead to similar and stable performance, which indicates that the proposed method is very robust. Accordingly, we set the radius parameter as 5 in the following experiment. For the interval $i$, the success plots AUC scores are listed in Table III, where the best result is in red. Table III illustrates that when $i = 1$, the proposed method with two frames outperforms the other parameters.

Table II
Success plots AUC scores of Vancouver data set search radius experiments

| Search radius | 4 | 5 | 6 | 7 | 8 | 10 |
|---|---|---|---|---|---|---|
| AUC | 0.857 | 0.861 | 0.860 | 0.860 | 0.859 | 0.859 |

Table III
Success plots AUC scores of Vancouver data set interval experiments

| Interval | 1 | 2 | 3 | 4 | 5 | 6 |
|---|---|---|---|---|---|---|
| AUC | 0.861 | 0.856 | 0.856 | 0.856 | 0.853 | 0.853 |

As we have determined the best pair of interval and search radius ($i = 1, r = 5$), parts of tracking results are shown in the Fig. 7. The Success plots and precision plots of Vancouver data set is shown in Fig. 8.

It can be observed in Fig. 7, that most state-of-the-art methods lost the target in the frames after 300 frames. Only the proposed method, CT, and Struck can track the moving train. And among them, the proposed method tracks the train more accurately. This can be proved in Fig. 8. As shown in Fig. 8, the proposed method gets a significant improvement than the other methods. In the success plot, our method outperforms the top ranked tracker of the comparison algorithm, Struck, with the AUC of 19.6% higher. In the precision plot, the proposed method also gets an AUC of 0.943, much higher than the second high AUC of 0.858. As Fig. 8 shown, in both success plots and precision plots, our method outperforms other state-of-the-art algorithms.

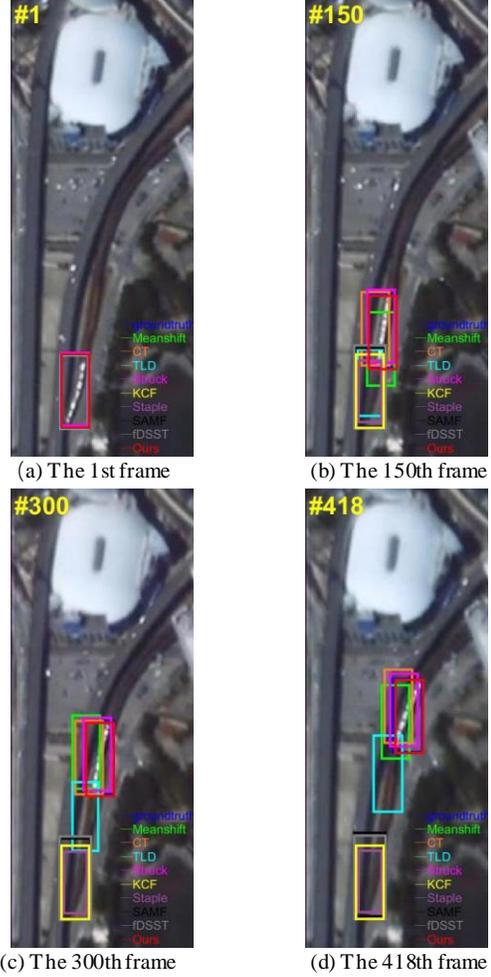

(a) The 1st frame  (b) The 150th frame
(c) The 300th frame  (d) The 418th frame
Fig. 7 Parts of the tracking results for Canada Vancouver harbor data set.

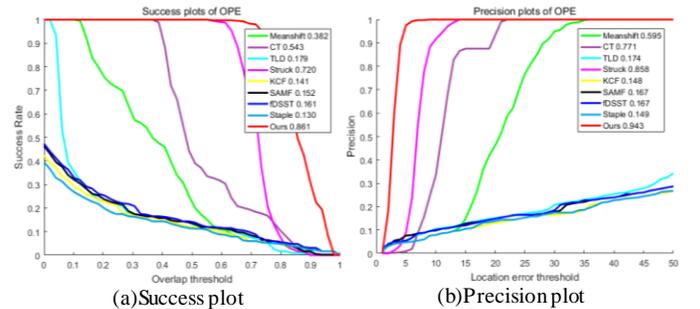

(a) Success plot  (b) Precision plot
Fig. 8 The Success plots and Precision plots of Vancouver data set. The performance score for each tracker is shown in the legend.

## B. India New Delhi

The second data set is an audio video interleaved (AVI) file provided by Chang Guang Satellite Technology Co.,Ltd.. The data set last 28 seconds, having 700 frames, and the frame size is $3600 \times 2700$, covering an urban area in New Delhi, India. The cropped area is from the original frame's coordinate (400,850), and the experimental area size is $650 \times 300$. The target we tracked is a train, where ground truth bounding box size is $72 \times 26$.

For the searching radius $r$, the success plot AUC scores are listed in Table IV, the best result is in red. Since the maximum moving distance of the train is 5 pixels, the search radius start from 5.

Table IV
Success plots AUC scores of New Delhi data set search radius experiments

| Search radius | 5 | 6 | 7 | 8 | 10 |
|---|---|---|---|---|---|
| AUC | 0.784 | 0.782 | 0.781 | 0.776 | 0.742 |

Table V
Success plots AUC scores of New Delhi data set interval experiments

| Interval | 1 | 2 | 3 | 4 | 5 | 6 |
|---|---|---|---|---|---|---|
| AUC | 0.784 | 0.803 | 0.812 | 0.809 | 0.818 | 0.814 |

According to Table IV, we set the radius parameter as 5. For the interval $i$, the success plot AUC scores are listed in Table V, the best result is in red.

According to the Table V, we can find that when the interval parameter $i = 5$, we can get the best result. Compared with last experiment, we can find that the movement of target between two neighboring frames in this experiment is much smaller. Since the target moves extremely slightly, the optical flow velocity between two neighboring frames would as approach to zero, thus the target is hard to be captured. As the interval between two frames become lager, the movement of the target between two frames will be more obvious, just like an accelerated movement of the target. Therefore, with the multi-frame difference method, the target of satellite videos will be better separated from the background. However, while the interval between two frames become too large, more background pixels will become distributing, and the optical field will become so messy that the accurate location of the interested target may be lost. Therefore, the interval parameter $i = 5$ shows a balance, where the optical flow velocity of the target is larger than that between two neighboring frames, and the background pixels still do not have too much velocity to interrupt the accurate location of the target.

As we have determined the best pair of interval and search radius ($i = 5, r = 5$), parts of tracking results are shown in the Fig. 9. The Success plots and precision plots of New Delhi data set are shown in Fig. 10. The performance score for each tracker is shown in the legend.

It can be observed in Fig. 9, that most state-of-the-art methods lost the target in the frames before the 200th frame. Only the proposed method, Meanshift, CT, and Struck can track the moving train. And among them, the proposed method tracks the train more accurately. And the remained algorithms also lost the target after the 500th frame, as shown in Fig. 9,. Only the proposed method has the ability to track the target during the whole tracking task.

As Fig. 10 shows, in both success plots and precision plots, our method outperforms other state-of-the-art algorithms obviously. The proposed method gets a significant improvement than the other state-of-the-art methods. In the success plot, our method gets an AUC of 0.818, outperforms the other top ranked tracker among the comparison algorithm, Struck, with the AUC of 0.245, threefold higher. In the precision plot, the proposed method also gets an AUC of 0.926, much higher than the second high AUC of 0.317.

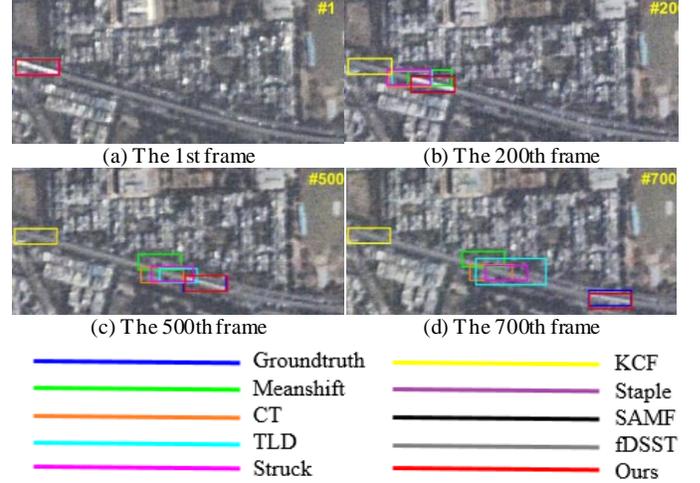

(a) The 1st frame    (b) The 200th frame

(c) The 500th frame    (d) The 700th frame

Groundtruth — KCF
Meanshift — Staple
CT — SAMF
TLD — fDSST
Struck — Ours

Fig. 9 Parts of the tracking results for India New Delhi harbor data set

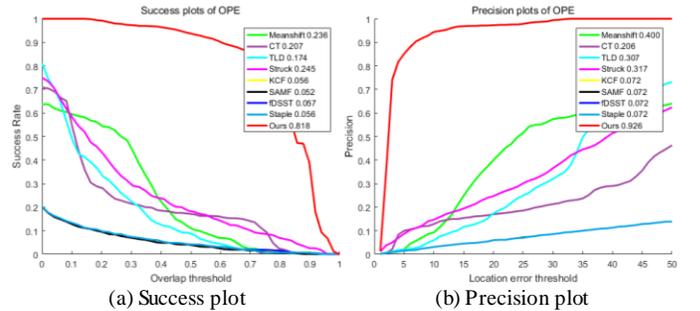

(a) Success plot    (b) Precision plot
Fig. 10 The Success plots and Precision plots of India New Delhi data set

## C. Afghanistan Kabul airport

The third data set is a MPEG-4 file acquired on Feb, 23th, 2017, provided by Chang Guang Satellite Technology Co.,Ltd.. The data set last 15 seconds having 375 frames, and the frame size is 3840×2160, covering an urban and airport area in Kabul, Afghanistan. The cropped area is from the original frame's coordinate (0,1860), and the experimental area size is 930×300. The target we tracked is a plane, where ground truth bounding box size is 17×15. Since after the 268th frame, the target flies out the range of video frame, thus we just employ the data set from the 1st frame to the 268th frame.

For the searching radius $r$, the success plot AUC scores are listed in Table VI, the best result is in red.

According to the Table VI, when the search radius parameter $r = 7$, the proposed method gets the best result. It is because in this experiment, the interested target is a plane. Since the plane moves much quicker than a train, the movement of the target between two neighboring frames will be much larger than that in previous two experiments. If the search radius is too small, the plane may move out the search area, and the interested target will be lost. While the search radius become too big, a larger search area will be considered, and some quickly moving background pixels will interrupt the calculation of the accurate location of the interested target. So, when the search radius $r = 7$, the search area includes the whole target candidate area, and not include too much interrupt background pixels.

Table VI
Success plots AUC scores of Kabul data set search radius experiments

| Search radius | 5 | 6 | 7 | 8 |
|---|---|---|---|---|
| AUC | 0.293 | 0.308 | 0.526 | 0.321 |

Table VII
Success plots AUC scores of Kabul data set interval experiments

| Interval | 1 | 2 | 3 | 4 |
|---|---|---|---|---|
| AUC | 0.526 | 0.521 | 0.518 | 0.386 |

According to Table VI, we set the radius parameter as 7. For the interval $i$, the success plot AUC scores are listed in Table VII, the best result is in red.

According to Table VII, we can find that the best result of interval parameter become $i = 1$ again. In this experiment, the interested target, the plane, moves quickly. Since the optical flow velocity between two neighboring frames has become large enough for separating target from the background, we do not need to take an interval between two frames

As we have determined the best pair of interval and search radius ($i = 1, r = 7$), parts of tracking results are shown in the Fig. 11. The success plots and precision plots of Afghanistan data set are shown in Fig. 12. The performance score for each tracker is shown in the legend.

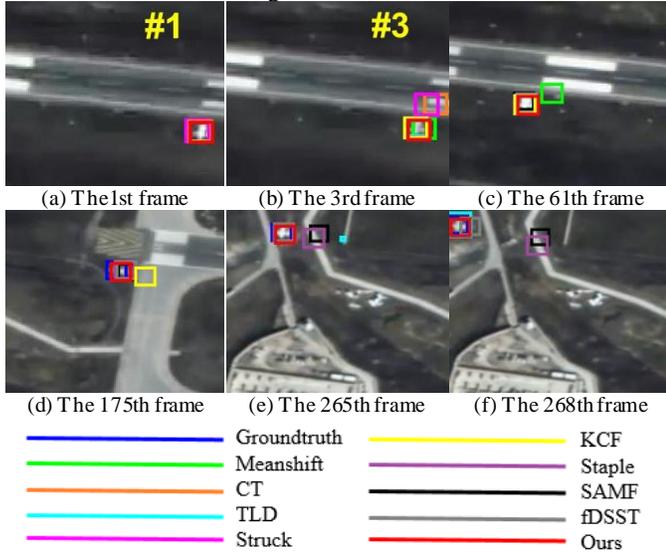

(a) The 1st frame (b) The 3rd frame (c) The 61th frame
(d) The 175th frame (e) The 265th frame (f) The 268th frame

— Groundtruth — KCF
— Meanshift — Staple
— CT — SAMF
— TLD — fDSST
— Struck — Ours

Fig. 11 Parts of tracking results of Afghanistan Kabul harbor data set.

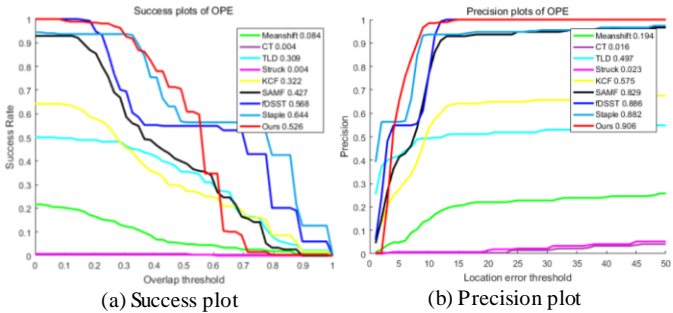

(a) Success plot (b) Precision plot

Fig. 12 The Success plots and Precision plots of Afghanistan Kabul data set

It can be observed in Fig. 11 (b) that Struck and CT algorithms lost the target after the third frame. Fig. 11 (c) shows that the Meanshift lost the target after the 61th frame. Fig. 11 (d) and Fig. 11 (e) show that since around the 175th frame and the 265th frame, where the cement road background is very similar to the interested plane, KCF, SAMF, Staple algorithm lost the target as well. Since TLD algorithm contains the full image scan, it lost the target frequently but captured the target some time, thus it can be seen in some frame but lost the target again soon. Only the proposed method and the fDSST algorithm can track the target during whole tracking task. It can be testified by Fig. 12, where only the proposed method and fDSST algorithm can obtain the 100% success rate when overlap threshold is less than 10% or location error threshold is larger than 10 pixels.

In the success plots in Fig. 12, our method's AUC score is the third best while compared with other state-of-the-art algorithms. It is worth noting that, our method gets an AUC of 0.906, outperforming the top ranked tracker of the comparison algorithm, fDSST, with the AUC of 0.886, by 2.3% in precision plots.

IV. CONCLUSION

Object tracking technology with satellite video data has significant potential in motion analysis [25], traffic monitoring [26, 27], suspicious object surveillance [29] el al. However, satellite video object tracking has not been studied widely and intensively, yet. In this paper, we proposed a novel object tracking method —Multi-Frame Optical Flow Tracker (MOFT), that aimed at tracking object in satellite video datasets. The proposed method first disposed frames to get optical flow field by optical flow method. Then, the HSV color converts the optical flow field into three bands colorful image. Finally, the integral image is employed to obtain the most probable position of target. Besides, the multi-frame difference method is employed to get a more accurate position of target.

Three experiments on VHR remote sensing satellite video datasets were employed for quantitative evaluation. The results indicate that the proposed method has the ability to track slightly moving object more accurately. Compared with other state-of-the-art algorithms, the proposed method provides a better way to track satellite video targets.

In the proposed method, the search radius and interval parameters are correlated with the move speed of the target. While the target moves slowly, like a slowly moving train, we need to increase the interval parameter to get a larger movement velocity between two frames, and the search radius can be smaller. On the other hand, while the target moves quickly, like a plane, we can increase the search radius to include the full candidate area of the interested target. Since the movement is quick enough, we do not need an interval between two frames, and using neighboring frames for tracking task is a better choice.

In our future work, we will focus on two aspects: 1) employ correlation filter to complete a better and efficient tracking task. 2) apply machine learning method to have a better tracking result.

ACKNOWLEDGMENT

This work was supported by the National Natural Science Foundation of China under Grants 61601333, 61471274, and by the China Postdoctoral Science Foundation under Grants 2016T90733. The authors would like to thank Deimos Imaging, UrtheCast, and Chang Guang Satellite Technology Co. Ltd for acquiring and providing the data used in this study, and the IEEE GRSS Image Analysis and Data Fusion Technical Committee.

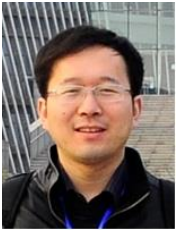

**Bo Du** (M'10–SM'15) received the B.S. degree and the Ph.D. degree in Photogrammetry and Remote Sensing from State Key Lab of Information Engineering in Surveying, Mapping and Remote sensing, Wuhan University, Wuhan, China, in 2005, and in 2010, respectively.

He is currently a professor with the School of Computer, Wuhan University, Wuhan, China. He has more than 40 research papers published in the IEEE Transactions on Geoscience and Remote Sensing (TGRS), IEEE Transactions on image processing (TIP), IEEE Journal of Selected Topics in Earth Observations and Applied Remote Sensing (JSTARS), and IEEE Geoscience and Remote Sensing Letters (GRSL), etc. Five of them are ESI hot papers or highly cited papers. His major research interests include pattern recognition, hyperspectral image processing, and signal processing.

He is currently a senior member of IEEE. He received the best reviewer awards from IEEE GRSS for his service to IEEE Journal of Selected Topics in Earth Observations and Applied Remote Sensing (JSTARS) in 2011 and ACM rising star awards for his academic progress in 2015. He was the Session Chair for both International Geoscience and Remote Sensing Symposium (IGARSS) 2016 and the 4th IEEE GRSS Workshop on Hyperspectral Image and Signal Processing: Evolution in Remote Sensing (WHISPERS). He also serves as a reviewer of 20 Science Citation Index (SCI) magazines including IEEE TGRS, TIP, JSTARS, and GRSL.

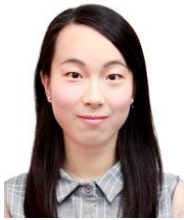

**Shihan Cai** received the B.S degree in International School of Software from Wuhan University, Wuhan, China, in 2017. She is currently pursuing the M.S degree in the school of computer, Wuhan University.

Her research interests include object tracking, and machine learning.

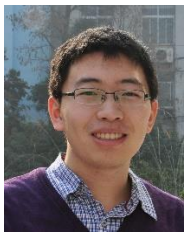

**Chen Wu** (M'16) received B.S. degree in surveying and mapping engineering from Southeast University, Nanjing, China, in 2010, and received the Ph.D. degree in Photogrammetry and Remote Sensing from State Key Lab of Information Engineering in Surveying, Mapping and Remote sensing, Wuhan University, Wuhan, China, in 2015.

He is currently a lecturer with the International School of Software, Wuhan University, Wuhan, China. His research interests include multitemporal remote sensing image change detection and analysis in multispectral and hyperspectral images.

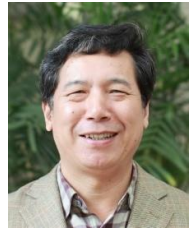

**Liangpei Zhang** (M'06–SM'08) received the B.S. degree in physics from Hunan Normal University, Changsha, China, in 1982, the M.S. degree in optics from the Xi'an Institute of Optics and Precision Mechanics, Chinese Academy of Sciences, Xi'an, China, in 1988, and the Ph.D. degree in photogrammetry and remote sensing from Wuhan University, Wuhan, China, in 1998.

He is currently the head of the remote sensing division, state key laboratory of information engineering in surveying, mapping, and remote sensing (LIESMARS), Wuhan University. He is also a "Chang-Jiang Scholar" chair professor appointed by the ministry of education of China. He is currently a principal scientist for the China state key basic research project (2011–2016) appointed by the ministry of national science and technology of China to lead the remote sensing program in China. He has more than 500 research papers and five books. He is the holder of 15 patents. His research interests include hyperspectral remote sensing, high-resolution remote sensing, image processing, and artificial intelligence.

Dr. Zhang is the founding chair of IEEE Geoscience and Remote Sensing Society (GRSS) Wuhan Chapter. He received the best reviewer awards from IEEE GRSS for his service to IEEE Journal of Selected Topics in Earth Observations and Applied Remote Sensing (JSTARS) in 2012 and IEEE Geoscience and Remote Sensing Letters (GRSL) in 2014. He was the General Chair for the 4th IEEE GRSS Workshop on Hyperspectral Image and Signal Processing: Evolution in Remote Sensing (WHISPERS) and the guest editor of JSTARS. His research teams won the top three prizes of the IEEE GRSS 2014 Data Fusion Contest, and his students have been selected as the winners or finalists of the IEEE International Geoscience and Remote Sensing Symposium (IGARSS) student paper contest in recent years.

Dr. Zhang is a Fellow of the Institution of Engineering and Technology (IET), executive member (board of governor) of the China national committee of international geosphere–biosphere programme, executive member of the China society of image and graphics, etc. He was a recipient of the 2010 best paper Boeing award and the 2013 best paper ERDAS award from the American society of photogrammetry and remote sensing (ASPRS). He regularly serves as a Co-chair of the series SPIE conferences on multispectral image processing and pattern recognition, conference on Asia remote sensing, and many other conferences. He edits several conference proceedings, issues, and geoinformatics symposiums. He also serves as an associate editor of the *International Journal of Ambient Computing and Intelligence, International Journal of Image and Graphics, International Journal of Digital*

*Multimedia Broadcasting, Journal of Geo-spatial Information Science, and Journal of Remote Sensing*, and the guest editor of *Journal of applied remote sensing* and *Journal of sensors*.

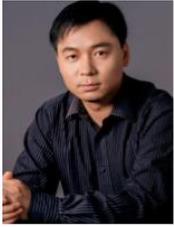

**Dacheng Tao**(F'15) is Professor of Computer Science and ARC Laureate Fellow in the School of Information Technologies and the Faculty of Engineering and Information Technologies, and the Inaugural Director of the UBTECH Sydney Artificial Intelligence Centre, at the University of Sydney. He mainly applies statistics and mathematics to Artificial Intelligence and Data Science. His research interests spread across computer vision, data science, image processing, machine learning, and video surveillance. His research results have expounded in one monograph and 500+ publications at prestigious journals and prominent conferences, such as IEEE T-PAMI, T-NNLS, T-IP, JMLR, IJCV, NIPS, ICML, CVPR, ICCV, ECCV, ICDM; and ACM SIGKDD, with several best paper awards, such as the best theory/algorithm paper runner up award in IEEE ICDM'07, the best student paper award in IEEE ICDM'13, the distinguished student paper award in the 2017 IJCAI, the 2014 ICDM 10-year highest-impact paper award, and the 2017 IEEE Signal Processing Society Best Paper Award. He received the 2015 Australian Scopus-Eureka Prize, the 2015 ACS Gold Disruptor Award and the 2015 UTS Vice-Chancellor's Medal for Exceptional Research. He is a Fellow of the IEEE, AAAS, OSA, IAPR and SPIE.